# Semantic Interoperability Middleware Architecture for Heterogeneous Environmental Data Sources

Adeyinka K. AKANBI, Muthoni MASINDE
*Department of Information Technology, Central University of Technology, Free State,*
*South Africa. Tel: +27 515073457, Email: aakanbi@cut.ac.za*
*South Africa. Tel: +27 515073091, Email: emasinde@cut.ac.za*

**Abstract:** Data heterogeneity hampers the effort to integrate and infer knowledge from vast heterogeneous data sources. An application case study is described, in which the objective was to semantically represent and integrate structured data from sensor devices with unstructured data in the form of local indigenous knowledge. However, the semantic representation of these heterogeneous data sources for environmental monitoring systems is not well supported yet. To combat the incompatibility issues, a dedicated semantic middleware solution is required. In this paper, we describe and evaluate a cross-domain middleware architecture that semantically integrates and generate inference from heterogeneous data sources. These use of semantic technology for predicting and forecasting complex environmental phenomenon will increase the degree of accuracy of environmental monitoring systems.

## 1. Introduction

The past half-century has witnessed rapid advancement in various areas of computing and Information Technology [1], with smart environments now representing the next evolutionary development step in the home and environmental monitoring systems. The notion of smart environment evolves from the definition of ubiquitous systems. According to [2] promotes the idea of "*a physical world that is richly and invisibly interwoven with sensors, actuators, displays, and computing element seamlessly in the everyday objects of our lives, and connected through a continuous network.*"

 Enabling technologies needed for the realization of the smart environment is multifaceted and most especially involves; wireless communication, algorithm design, multi-layered software architecture (*middleware*), event processing engines, sensors, semantic web, knowledge graphs, adaptive control amongst others. Currently, the integration of all these technologies is at their infant stages; mostly due to heterogeneity in ubiquitous components. The expectation that networks of heterogeneous smart devices and services can be integrated to form an information system is driving the need for broad agreement on data integration and interoperability across software boundaries.

 However, in environmental monitoring domain, forecasting complex environmental phenomenon such as drought involves combining diverse data sources (*e.g., sensor data, weather station data, geospatial data, satellite imagery, indigenous knowledge*) for accurate forecasting information. Eventually, this leads to processing and integration of a large amount of heterogeneous data from multiple sources. Because of the complexity of the environmental phenomenon and heterogeneity of the input data sources, a supporting information system for the integration is required. Unfortunately, data integration and interoperability is a universal challenge and demanding task that have an effect on all spheres of information systems.

On one hand, resources/things such as sensor devices are manufactured by different vendors using different data representation format and communication protocol. On the other hand, unstructured knowledge base like the local indigenous knowledge needs to be augmented with data from computational models. In weather monitoring/forecasting field, indigenous knowledge reflects unique expertise in the local environment [15]. Either way, these data sources are heterogeneous and has to be integrated effectively. Hence, this work focuses on the seamless integration and exchange of data among these resources/things for drought forecasting; and how to facilitate efficient interoperability among different entities by solving the issue of data heterogeneity which is a major bottleneck for the realization of accurate environmental monitoring systems.

According to [2], interoperability is "*the ability of two or more systems or components to exchange data and use information.*" We believe that in order to realize the anticipated integration of these heterogeneous data sources there is a need for high-level interoperability. However, there are several levels of interoperability ranging from technical, syntactic, semantic, pragmatic interoperability and organizational interoperability [3,14]. In order to achieve data integration and interoperability in environmental monitoring domain, semantic level interoperability offers the technologies needed for enabling the same meaning to an exchange piece of data to be shared by communicating nodes. It achieves this through the representation of the data in a machine-readable format using knowledge description and automated reasoning.

Moreover, modern sensory and legacy devices communication systems were open systems built using the manufacturer's unique data and communication standards and thus requires common semantic level interoperability solutions. Consequently, there has been wide adoption of semantic technologies in smart information systems and Internet of Things related projects. For example, Networked Embedded System Middleware for Heterogeneous Physical Devices (HYDRA)[1], Internet Connected Objects for Reconfigurable Ecosystems (iCore)[2], Context Broker Architecture for Pervasive Computing (CoBrA) [4] to mention a few.

In this regard, we propose a functional semantic data integration and interoperability middleware architecture for heterogeneous environmental data sources. The architecture is composed of several components all implemented using semantic technologies. As a whole, the paper makes the following contributions:

- A functional semantics-based heterogeneous data integration architecture is going to be introduced.
- An architecture for collecting data from sensor devices and local indigenous knowledge on drought is proposed to facilitate seamless data integration and interoperability.
- Ontology designed for the formalization of the Indigenous knowledge on drought
- System architecture process flowchart.

The rest of the paper is organized as follows: in Section 2 we introduce the background; Section 3 describes the architecture of the middleware. Section 4 outlay the case study and Section 5 we present the result. In Section 6 we conclude this work and outline future work.

---

[1] See http://projecthydra.org/
[2] See http://www.iot-icore.eu/

## 2. Background (State of The Art)

Advancement in the field of wireless sensor network (WSN) and the Internet of Things (IoT) has facilitated the use of sensors in the collection of accurate sensory data for environmental monitoring. The data comes from multiple sensors of different modalities in distributed locations making up the WSN as well as legacy systems. The heterogeneity of the data sources, however, is a major challenge towards the integration of these data sources gives rise to the problem of compatibility, interoperability, integration of services, and difficulty in generating environmental inference from the data. The difficulty of achieving this seamless integration and full interoperability of interconnected heterogeneous devices is, however, due to the heterogeneity of the ubiquitous components, and data formats.

Traditionally, various effort to address interoperability is to define standards as been done many individual manufacturers [18]. There are however many standards and specifications that are incompatible with each other. The promising technology to tackle these problems of interoperability and integration of ubiquitously interconnected objects is the semantic technologies [2]. Semantic technologies have a stronger approach to interoperability than contemporary standards-based approaches [3], such as Service Oriented Architecture (SOA) through detailed semantic referencing of metadata.

## 3. Service Oriented Architecture of Heterogeneous Data Integration Middleware

This section outlines the technical infrastructure for the proposed data integration middleware and the required technology. We adopted a layered Service Oriented Architecture (*SOA*) in which each layer encompasses components (*functional groups*). Each functional groups (*FG*) is composed of several modules and offers high level of and provides functionalities suitable to the level. One of the most remarkable features of the proposed middleware infrastructure is the provision of an inference engine component. The adoption of the inference engine and related technologies allows the generation of accurate inference from the data sources. Also, the middleware acts like a bond joining heterogeneous domains of application community over heterogeneous interfaces. It also provides Application Programming Interface (*API*) for physical layer communication, abstraction of complex network communication and presenting the data in a machine-readable format for easy usage and interoperability [3].

*3.1 System Architecture Overview*

The main fundamental characteristic of the proposed semantic level interoperability middleware is the ability to integrate both structured data (*sensors data*) and unstructured data (*indigenous knowledge*). The middleware is novel and revolutionary for semantically integrating all heterogeneous data sources such as the weather station data, sensory data from the wireless sensors, and the indigenous knowledge of the local people on drought. The proposed architecture is composed of four functionality groups (*FG*): *Data Acquisition FG*, *Data Storage FG*, *Stream Analytics FG*, *Inference Engine FG* and the *Data Publishing FG*. Figure 1 shows the architecture of the middleware. Each of these *FGs* is loosely coupled to allow them to be deployed across different servers. Therefore, in order to handle such huge amount of data from heterogeneous data sources, the middleware process the data in event fashion manner. Furthermore, the middleware encapsulates a set of domain models to identify the entities/resources from the unstructured data sources using the developed IKON Domain Ontology in the Inference Engine FG. The data are processed by through the *Stream Analytics FG* and the *Inference Engine FG*, where a deductive inference is conducted based on system rules by the reasoners. The inferred data is published across multiple channels by the *Data Publishing FG*. The RESTful[3] services are developed to make communication between the middleware functional groups; detailed information about the functional groups are presented in the following sections.

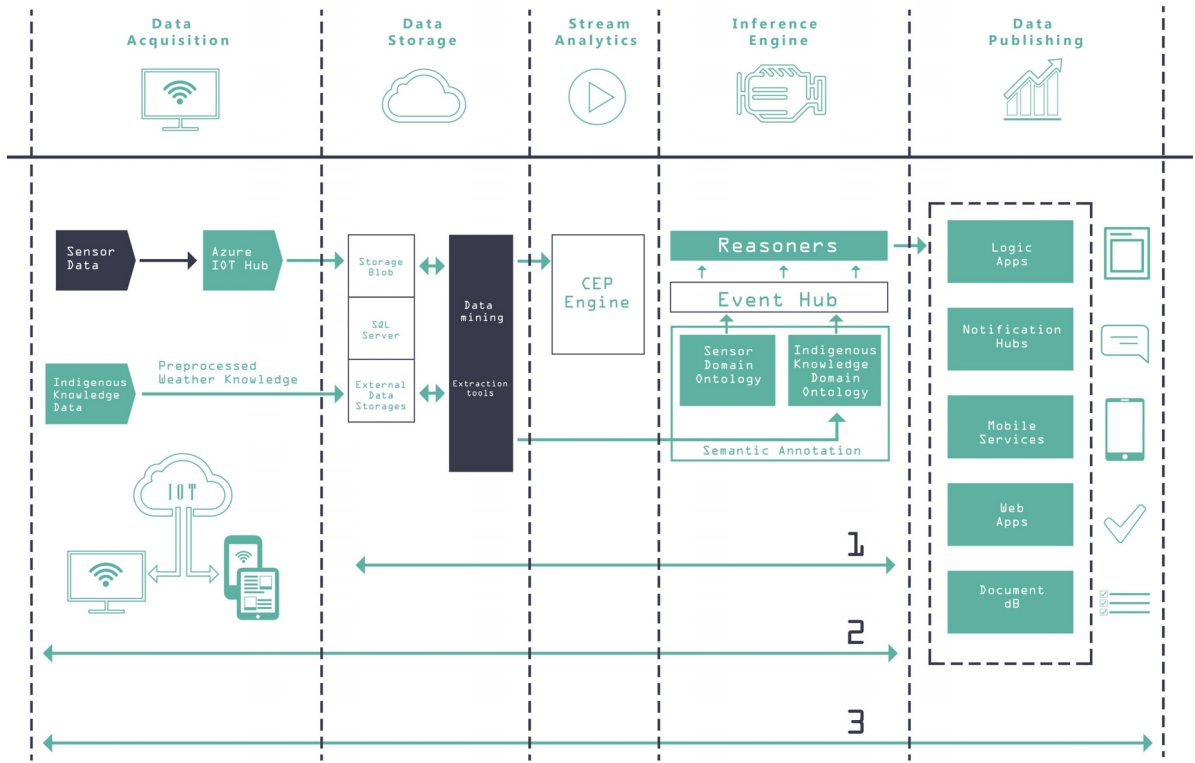

*Figure 1: An overview of the semantics-based data integration middleware architecture.*

### 3.2 Data Acquisition Functional Group

The Data Acquisition FG encapsulate the sensor data collection and the indigenous knowledge system modules of the architecture. This FG provides the necessary tools and technologies for data collection from different heterogeneous sources.

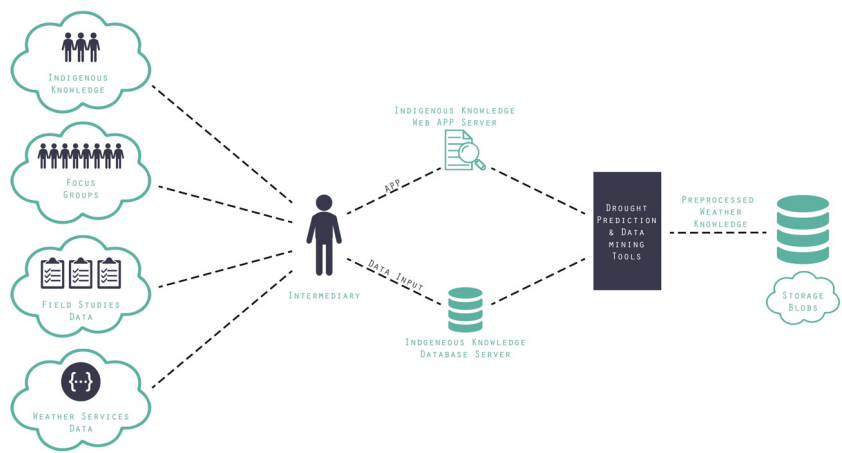

*Figure 2: Architecture of the Indigenous Knowledge System (IKS) module.*

---

[3] RESTful web services provide interoperability between computer systems on the Internet [19].

 

The Indigenous Knowledge System (IKS) module of the *Data Acquisition FG* as depicted in Figure 2 provides an abstraction for collecting, gathering, and documentation of the local indigenous knowledge. The local indigenous knowledge is paramount to the realization of the system because it offers the desired level of scalability and granularity of use of indigenous knowledge systems in the area of context. Indigenous knowledge is oral, tacit, scattered and unstructured knowledge being used by indigenous people in certain geographic location. IK data were obtained from the domain experts or focus groups, through a series of oral consultation, interviews, field studies and meeting sessions through an intermediary. The unstructured data (*IK*) are temporarily stored in the indigenous knowledge database server, besides, an Android application developed for this research captures the type (pictorially), description and spatial coordinates of identified natural indicators remotely and stored in the Indigenous Knowledge Web App Server, the web services interact with the database using the standard JDBC connection. The data captured knowledge base are processed by the data mining tools into a form that is stored in the *Data Storage FG*.

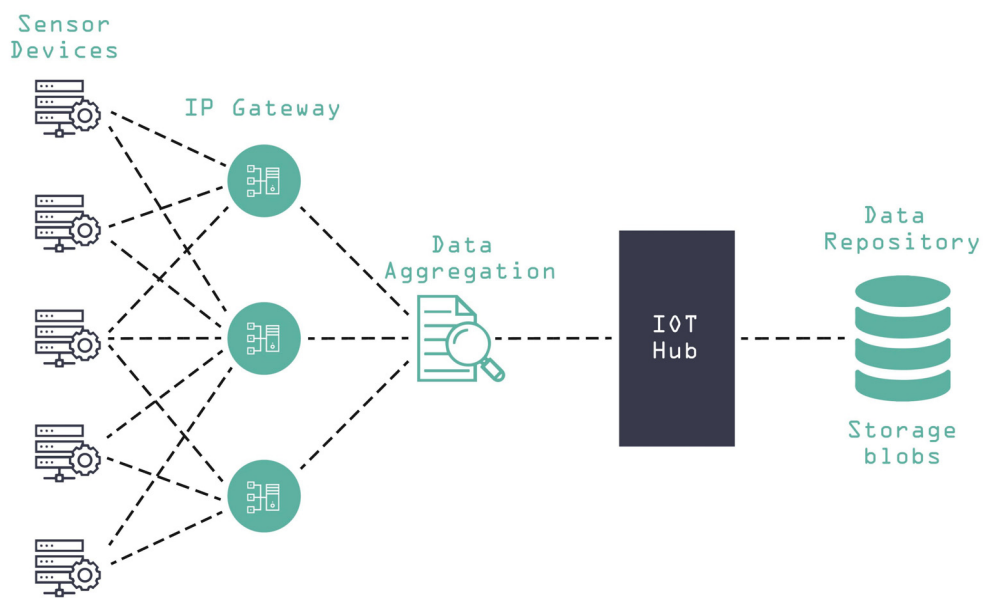

*Figure 3: Architecture of the Sensor Data Collection (SDC) module.*

The sensor data collection (SDC) module provides the overview to a network of connected sensors devices. The WSN (*sensors, boards, interface, radio transmitter*) will be set up to provide an accurate environmental data such as soil moisture, crop yield, leaf index and vegetable growth. The data are transmitted to the IoT hub (*e.g. Azure IoT[4], Google Cloud[5], Apache Kafka[6]*) via the gateway using designated communication medium, mostly based on 6LoWPAN protocol which enables transmission of compressed IPv6 packets over IEEE 802.15.4 networks.

---

[4] See https://azure.microsoft.com/en-us/suites/iot-suite/
[5] See https://cloud.google.com/
[6] See https://kafka.apache.org/

### 3.3 Data Storage Functional Group

Data received from the input data sources are of different types and formats (structured and unstructured). The *Data Storage FG* comprised of different types of modules that will facilitate the storage and processing of the raw data. Preliminarily, the raw data are transferred to this module for storage in the storage blobs, SQL Server or external storage mediums. From the data storage, various data mining and extraction tools are used to present the data in a form that can be processed by the *Stream Analytics FG*, *Inference Engine FG*.

### 3.4 Stream Analytics Functional Group

The sequence of sensor data from the sensors is processed from the cloud using a complex event processing (*CEP*) engine. The CEP engine infers patterns from the series of sensor data from the environmental monitoring system [5, 13]. The CEP engine detects composite events — from the observation of specific patterns of the sensor data. The ability of the CEP engine to infer the pattern of the event is achieved through CEP rules[7] that are embedded part of the application logic [6]. In this context, rules are in this general syntax:

*CE ($A_1 = J_1$ (..) , ..., $A_n = J_n$ (...) ) := Pattern*

Where the symbol := separates the rule head from the pattern. *CE* specifies the composite event captured by the rule and how its attributes $A_1,...,A_n$ are functionally defined by the attributes of the events that appear in the pattern. When a pattern is detected within the stream of input sensor data. The CEP engine knows that the corresponding composite event has occurred based on the specified CEP rule, and notifies the interested components if the stream of input events satisfies the pattern [6]. For example, data from four sensors $S_1...S_4$ will serve as input to the CEP engine in form of $S_1:=A_1 (T_1)$. The attribute value for the sensor is captured as well as the corresponding time stamp. A temperature sensor can capture four different reading within an hour period. Based on the CEP rule the average of those reading can trigger a pattern and used to infer an event such as '*High Temp.*' The inferred information is passed to the next *Inference Engine FG*.

### 3.5 Inference Engine Functional Group

The *Inference Engine FG* comprises of the Sensor Domain Ontology module, Indigenous knowledge Domain Ontology (IKON) module, Event Hub, and the Reasoners. This section of the middleware represents the data according to the domain models and performs deductive inference from the Event Hub and Reasoners. The ontology modules provide semantic metadata for the representation of knowledge about the given domain and provide core resources for reasoning and inference for the domain [7,12,17]. The domain ontologies in the *Inference Engine FG* addresses the need of a uniform representation for the data (*structured and unstructured*) in a way to be understood and processed by the reasoning engine module and support SPARQL queries.

In our research, for the sensor data input data, we adopted the Semantic Sensor Ontology [8] ontology which describes the capabilities and properties of sensors and the resulting observation. Figure 4 shows the relations of classes and properties of the ontology. Pre-processed data from the cloud are represented in XML, which represents the document structure (syntax) but lacks the explicit meaning (semantics) of the content. The XML documents are converted into a machine-readable language - OWL-Description Language

---

[7] A CEP rule defines a composite events from a pattern of events [6].

(OWL-DL) and aligned to the SSN ontology for the reasoning engine. After the process of ontology alignment that data can be onterpreted and processed by the reasoner.

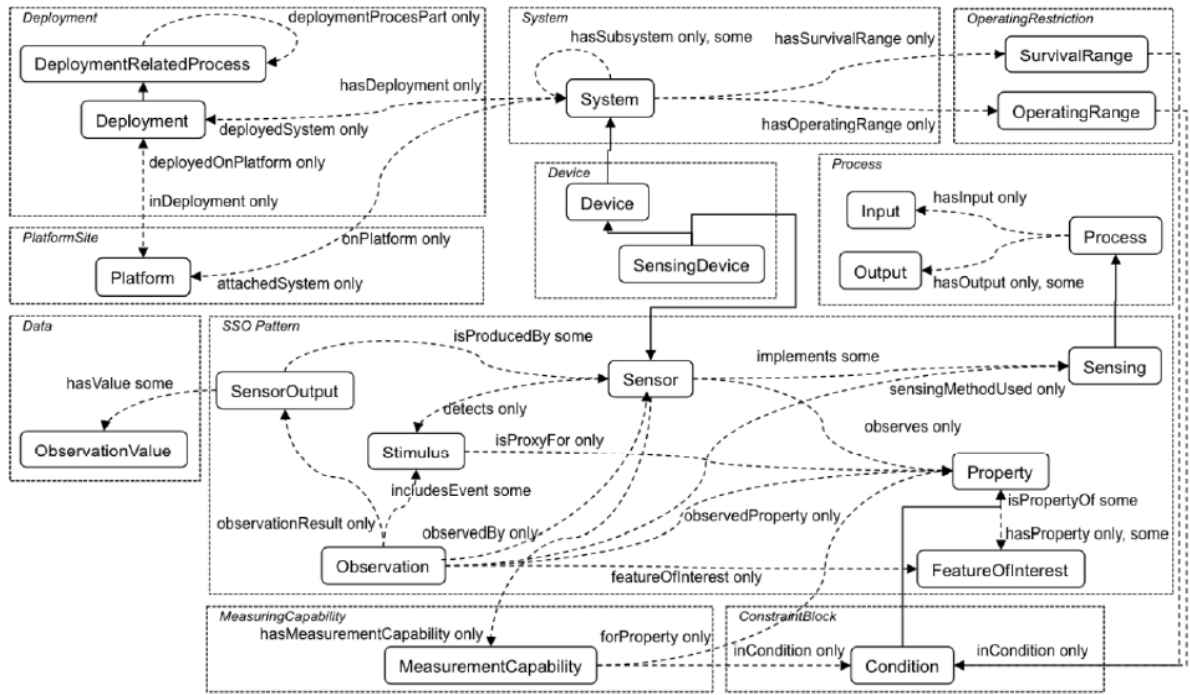

*Figure 4: Semantic Sensor Network Ontology (Adapted from [8]).*

For the unstructured indigenous knowledge (IK) we had a major challenge in front of us due to the lack of domain ontology that explicitly represents the local indigenous knowledge on drought. Therefore, our research group[8] developed from the scratch a domain ontology called Indigenous Knowledge Domain Ontology (IKON)[9]. It is designed to semantically represent the entities and event (*behavioral/observation*) in the indigenous knowledge domain using a minimal number of classes, properties and restrictions. Figure 5 depicts the hierarchical representation of the Indigenous knowledge on Drought Ontology (IKON). The five main classes were classified under the *owl:Things* into superclasses, *owl:LivingThings*, *owl:NonLivingThings*, *owl:LivingThingsBehaviour*, o*wl:NonLivingThingsBehaviour* and *owl:Event*. Each of them with their own hierarchy of subclasses. Based on the expert knowledge, the domain was classified and the mapping of the domain classes to the ontology was achieved through object-oriented techniques using multiple inheritances. The IK domain ontology does not aim at representing all types of events or scenarios. It provides a set of concepts and relations to model the events or scenario in the domain.

The event hub and semantic reasoners can infer logical patterns from a set of input data, anticipate an event and detect similarities. Applying formal representation to all data using ontology ensures all data exchanged in the *Inference Engine FG* are encoded as Ontology Web Language (OWL) documents for inference generation.

---

[8] See www.africrid.com
[9] Ontology publicly available at http://github.com/yinchar/Indigenous-Knowledge-Domain-Ontology.git and http://www.semanticweb.org/aakanbi/ontologies/2016/0/IKON

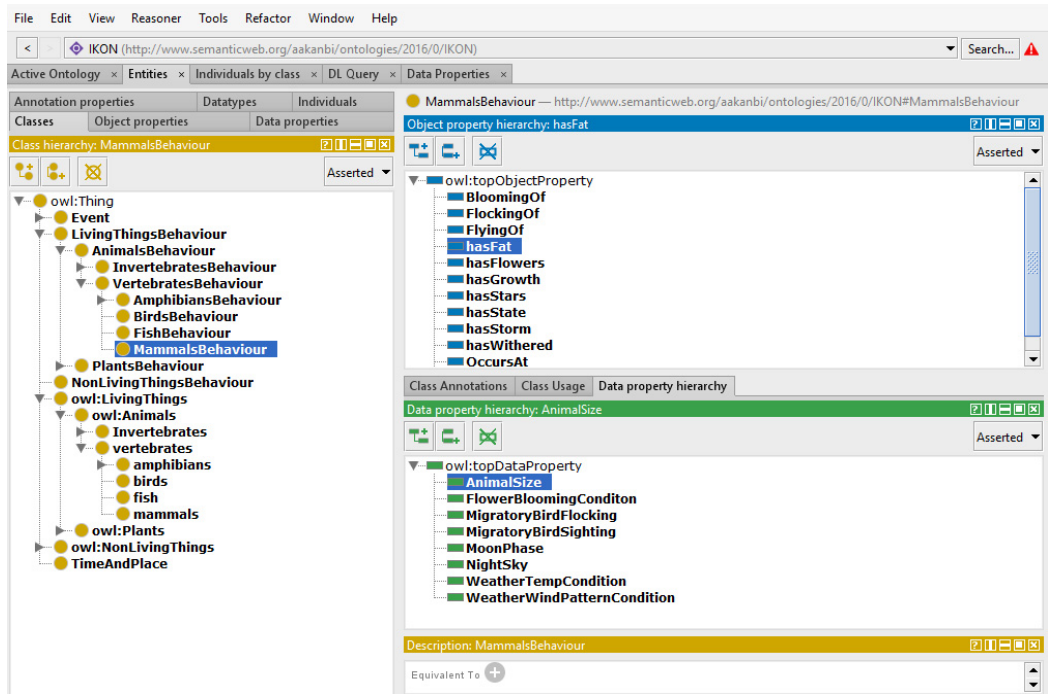

*Figure 5: The hierarchical representation of the Indigenous knowledge on Drought Ontology (IKON) with defined object and data properties.*

*3.6– Data Publishing Functional Group*

This *functional group* encompasses several modules for the distribution and publishing of the accurate deduced information across multiple channels. The information is in form of recommendation and has attributed certainty factor (CF) to indicate the level of confidence in the output information. This recommended information is in a form that is machine-readable, promote reuse and allows integration with other third-party applications using REST APIs.

## 4  Case Study: Semantics-Based Drought Early Warning System

The section introduces an application case study – Semantics-based Drought Early Warning System - DEWS around which we are proposing the architecture and implement a prototype. The main goal of the middleware is to hide the complexity and heterogeneity of environmental data sources [9]. The middleware integrates heterogeneous data sources such as modelled sensor data and local indigenous knowledge. To illustrate our proposal, we use as an example a drought forecasting system. Most environmental monitoring systems rely on environmental parameters such as the temperature, humidity, evapotranspiration measured in the area under study [16]. The measured environmental parameters serve as input data for different environmental computational analysis and modelling. However, due to the complexity of environmental phenomena such as drought and the variability of the measured environmental parameters. Researchers [10] are proposing the semantic integration of sensor-based modelled data with local indigenous knowledge of the area under study towards a more accurate environmental forecasting information. The use of the indigenous knowledge with modern methods provides the necessary data to ensure the scalability of the inferred forecasting information.

However, the major challenge towards the actualization of this novel integration idea is the formalization of the indigenous knowledge, considering the fact that the knowledge is unstructured and full of ambiguities. A semantics-based data integration middleware will address and solve these challenges. Most especially the middleware should facilitate the semantic representation and integration of the heterogeneous data using domain ontologies. Hence, our proposed middleware comprises of five functional groups due to the nature of the challenges. Each *functional group* is made up of several modules for seamless integration and interoperability.

For the sensor-based data, a variety of sensors, such as rainfall sensor, humidity sensor or temperature sensor can be used to measure the environmental parameters. The sensor data are modelled using the SSN ontology for formalizing, reasoning and querying the knowledge base. In a similar context, developed domain ontology IKON is used for the semantic representation of the indigenous knowledge. In the reasoning process, the metadata annotated heterogeneous data are correlated and semantic correlation is established by the reasoners. For example, a meteorological drought recommendation information will be inferred if the following system rules are true, see code snippet below.

```
Metadata:
RULESTART "Domain==INDIGENOUS KNOWLEDGE"
    IF (MugumoTree is Flowering AND
        MoonSize is Full
        InyosiBees are hiding
        MigratoryBird is Flocking
        AverageDailyTemp is Low
        CloudCover is High
        WindSpeed is Low
        Relative Humidity is High
        Evotranspiration is High)
    THEN New Inference => Drought [METEOROLOGICAL && AGRICULTURAL && HYDROLOGICAL] 10%
```

*Figure 6: Sample of Reasoner System Rule.*

## 5. Result

We contributed a conceptual semantic middleware architecture and depicted the information exchange between the functional groups. The middleware architecture developed allows the seamless integration of heterogeneous data sources for the drought forecasting domain. The system allows formalism of data sources with standardized, precisely defined syntax and formal semantics.

## 6. Conclusion and Future Work

Data integration is a universal challenge. This paper presents a vision of how integration and interoperability of heterogeneous data sources can be achieved for the drought forecasting and environmental monitoring systems. Case studies are also included. The middleware architecture proposed acts as the main catalyst for data integration providing the contrivance for the semantic data representation, annotation, generation of inference and reasoning. The system generates levels of forecasting recommendation with attributed certainty factors based on the input data from the sensors and the local indigenous knowledge for the end users which are the farmers. The overall solution on drought is based on the ITIKI framework for drought prediction [11]. This middleware takes processing, representation and dissemination of drought forecasting data from the current Web 2.0 to the Web 3.0, where information will be shared in a machine-readable format for effective environmental monitoring or forecasting in the realm of this latest technology. The overall middleware system is partly working and currently operates in test mode. In the future, we are planning to evaluate the feasibility, usability and acceptance of the middleware architecture and attract semantic application developer to benefit from the middleware architecture. Our proposed model can be enhanced to provide other levels of interoperability among heterogeneous environmental data sources.